# A Novel Energy Aware Node Clustering Algorithm for Wireless Sensor Networks Using a Modified Artificial Fish Swarm Algorithm

Reza Azizi[1], Hasan Sedghi[2], Hamid Shoja[3], Alireza Sepas-Moghaddam[4]

[1]Young Researchers and Elite Club, Bojnourd Branch, Islamic Azad University, Bojnourd, Iran

[2]Department of Information Technology Engineering, PNU, Assaluye, Iran

[3]Department of Computer Engineering and Information Technology, PNU, Tehran, Iran

[4]Young Researchers and Elite Club, Science and Research Branch, Islamic Azad University, Tehran, Iran

**ABSTRACT**

*Clustering problems are considered amongst the prominent challenges in statistics and computational science. Clustering of nodes in wireless sensor networks which is used to prolong the life-time of networks is one of the difficult tasks of clustering procedure. In order to perform nodes' clustering, a number of nodes are determined as cluster heads and other ones are joined to one of these heads, based on different criteria e.g. Euclidean distance. So far, different approaches have been proposed for this process, where swarm and evolutionary algorithms contribute in this regard. In this study, a novel algorithm is proposed based on Artificial Fish Swarm Algorithm (AFSA) for clustering procedure. In the proposed method, the performance of the standard AFSA is improved by increasing balance between local and global searches. Furthermore, a new mechanism has been added to the base algorithm for improving convergence speed in clustering problems. Performance of the proposed technique is compared to a number of state-of-the-art techniques in this field and the outcomes indicate the supremacy of the proposed technique.*

**KEYWORDS**

*Wireless Sensor Networks, Artificial Fish Swarm Algorithm, Node Clustering, Energy*

## 1.INTRODUCTION

Optimization technique is used in order to minimize or maximize the outcome of a function by adjusting its factors. All proper values for this problem are called possible solutions and the best one is optimal solution. Up to now, different approaches were proposed to perform optimization process e.g. swarm intelligence methods [1]. Swarm intelligence techniques are based on relations between animals and insects swarms which tries to detect the solutions that are close to optimal one. Particle Swarm Optimization (PSO) [2], Ant Colony Optimization (ACO) [3], Shuffled Frog Leaping Algorithm (SFLA) [4] and Artificial Fish Swarm Algorithm (AFSA) [5] are some examples of swarm intelligence algorithms. All swarm intelligence algorithms are based on population, where their iterative procedure leads to improve the position of individual in population and subsequently, their movement toward the better positions.

As mentioned above, Artificial Fish Swarm Algorithm (AFSA) is one of the swarm intelligence techniques [5]. The procedure of AFSA has been inspired from social behaviors of fish in nature, based on random search, population and behaviorism. This algorithm involves some characteristics i.e. high convergence speed, insensitivity to the initial values of artificial fish, flexibility, and fault tolerant which make it useful for solving optimization problems. Fish can find the zones involving more foods in underwater world by individual or swarm searches. According to this property, AFSA model was proposed by Free-Movement, Food-Search, Swarm-Movement and Follow Behaviors in order to search the problem space. This algorithm is used in different applications [6] such as neural networks [7, 8], color quantization [9], dynamic optimization problems [10], physics [11], global optimization [12-15], and data clustering [16].

In recent years, Wireless Sensor Network (WSN) has been significantly considered by researchers. Each sensor network includes a number of nodes, where each node involves processing resources and a limited amount of energy. Sensor nodes can sense, measure and collect information by some local decision and transmit it to the main station [17]. Intelligent sensor node consist of one or several sensors, one processor, a memory unit, a type of energy supply with low energy utilization, transmitter, receiver, and a stimulator. Battery is the primary source in a sensor node which is not commonly rechargeable, so the node will be dead by finishing batteries' energy. Network lifetime is determined by two criteria: 1) the time in which the first node dies (FND) and 2) the time in which the last node dies (LND). Utilizing different mechanisms to increase the lifetime of the network is considered amongst most challengeable subjects in this domain. Many researches aim at enhancing the network lifetime by proposing duty cycling schemes [18], coverage targeted protocols [19] or novel routing protocols [20].

One of the efficient solutions to increase the network's lifetime is clustering of the nodes [21]. Clustering is especially important for sensor networks which contain a substantial number of wireless sensors. Each network can be divided into smaller clusters and contain a cluster head (CH). Sensor nodes in every cluster transfer their information to the relevant CH and CH collects information and sends them to a primary station. In this way, cluster nodes can conserve more energy by reducing the range and length of the communication. Swarm and evolutionary algorithms can be used in the primary station to determine the members of each cluster along with the cluster heads.

In this research, a novel algorithm for clustering of nodes in wireless sensor networks has been proposed, based on AFSA. The proposed algorithm along with Leach algorithm [22] and several other ones have been used for the clustering process. The results of the simulations showed that the network's lifetime has been increased by using the proposed method, compared to other ones. The rest of the paper is organized as follow: in the second section Artificial Fish Swarm Algorithm is reviewed. The third section describes the proposed method and the forth one shows the experiments and the results. The fifth section concludes the paper.

## 2. ARTIFICIAL FISH SWARM ALGORITHM (AFSA)

AFSA is one of the swarm intelligence methods and evolutionary optimization techniques [5]. The framework of this algorithm is based on functions that are modelled from social interactions of fish groups in the nature.

In this algorithm, there are four functions modelled from fish behaviors in fish swarms. The first function is free-move behavior: as in nature, fishes move freely in the swarm when they do not prey. The second function is prey behavior: certainly, every fish searches for its prey individually by means of its senses. These senses are sense of vision, smell and available sensors on their bodies. In AFSA, the area where an artificial fish can sense a prey is modeled as a neighborhood with a visual-sized radius. The next function is follow behavior: when a fish finds food, other swarm members also follow it to reach the food. The last function is swarm behavior: in nature, fish always try to be in the swarm and not to leave it in order to be protected from hunters.

In the underwater world, a fish can discover areas that have more food, which is done by individual or swarm search by the fish. According to this characteristic, an artificial fish (AF) model is expressed by four aforementioned behaviors. AFs search the problem space by these behaviors. The environment, where an AF lives, is basically the solution space and other AFs domain. Food consistency degree in the water area is AFSA objective function. Finally, AFs reach to a point where its food consistency degree is maxima (global optimum).

As it is observed in Fig. 1, AF perceives external concepts with a sense of sight. The current position of an AF is shown by vector $X=(x_1, x_2, …,x_n)$. The visual is equal to sight field of the AF and $X_v$ is a position in visual where the AF wants to go. Then if $X_v$ has better food consistency than the current position of AF, it goes one step toward $X_v$ which causes change in the AF position from X to $X_{next}$, but if the current position of AF is better than $X_v$, it continues searching in its visual area. Food consistency in position X is fitness value of this position and is shown with f(X). The step is equal to the maximum length of the movement. The distance between two AFs which are in $X_i$ and $X_j$ positions is shown by $Dis_{i,j}=\| X_i-X_j \|$ (Euclidean distance) that is computed by Eq (1):

$$Dis_{i,j} = \sqrt[2]{\sum_{d=1}^{D}(X_{j,d} - X_{i,d})^2} \quad (1)$$

AF model consists of two parts of variables and functions. Variables include X (current AF position), step (maximum length step), visual (sight field), try-number (the maximum test interactions and tries), bulletin and crowd factor δ (0<δ<1). Also functions consist of prey behavior, free-move behavior, swarm behavior and follow behavior.

In each step of the optimization process, AF attempts to find locations with better fitness values in the problem search space by performing these four behaviors based on the algorithm procedure. In the following, the behaviors of AFSA will be discussed.

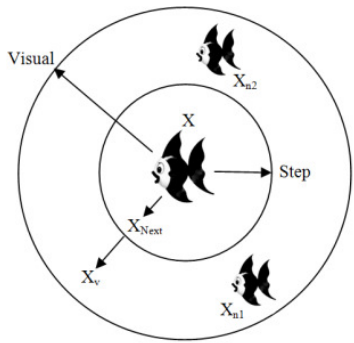

Figure 1. Artificial fish and the environment around it.

**Free-move behavior:** In AFSA, when an AF can't move toward a place with more food, it moves a arbitrary step in the problem space by Eq (2):

$$X_{i,d}(t+1) = X_{i,d}(t) + Step \times Rand_d(-1,1) \quad (2)$$

$X_{i,d}$ is component d of AF i's position in the D-dimensional space and $1 \leq d \leq D$. Rand function generates a random number with a uniform distribution in [-1, 1].

**Prey behavior:** If $X_i$ is the current position of AF i, we choose position $X_j$ in the visual of AF i randomly. f(X) is the food consistency in position X or its fitness value. Position $X_j$ is given by Eq (3):

$$X_{j,d} = X_{i,d} + Visual \times Rand_d(-1,1) \quad (3)$$

Then food density in $X_i$ is compared with the that of the current position, if $f(X_i) \geq f(X_j)$, AF i moves forward a step from its current position to $X_j$, which is done by Eq (4):

$$\vec{X}_i(t+1) = \vec{X}_i(t) + \frac{\vec{X}_j - \vec{X}_i(t)}{Dis_{i,j}} \times Step \times Rand(0,1) \quad (4)$$

$X_i$ is a D-dimensional vector and $Dis_{i,j}$ is the Euclidean distance between vectors $X_i$ and $X_j$. $X_j - X_i$ generates a transfer vector from $X_i$ to $X_j$ and when divided by $Dis_{i,j}$, a vector with unit length is created from $X_i$ toward $X_j$. Here, Rand function generates a random number which causes AF i to move as much as a random percent of Step toward position $X_j$. Nevertheless, if $f(X_i) < f(X_j)$, we choose another position $X_i$ by Eq (3) and evaluate its food density to understand whether forward condition is satisfied or not. If after try_number times AF does not succeed in satisfying forward condition, the concerned AF performs free-move behavior and moves a step in the problem space randomly.

**Swarm behavior:** In AFSA, in order to keep swarm generality, in each of the iterations, AFs try to move toward a central position. A central position of swarm is given by Eq (5):

$$X_{Center,d} = \frac{1}{N} \sum_{i=1}^{N} X_{i,d} \quad (5)$$

As it is seen in Eq (5), component d of $X_{center}$ vector is the arithmetic average of component d of all swarm AFs. Let N be the population size, and nc be the number of AF in Visual field around $X_{center}$. If $f(X_{center}) \leq f(X_i)$ and $\delta > (nc/N)$, that is the central position has a better food consistency than the current position and population density in its neighborhood is not much, so AF i moves toward the central position by Eq (6):

$$\vec{X}_i(t+1) = \vec{X}_i(t) + \frac{\vec{X}_{Center} - \vec{X}_i(t)}{Dis_{i,Center}} \times Step \times Rand(0,1) \quad (6)$$

If nc = 0 or the condition of moving toward the central position is not satisfied, prey behavior is performed for AF i.

**Follow behavior:** If Xi is the current position of AF i, it checks neighbor Xn, if nn is the number of AFs in the Visual of AF n, if $f(Xn) \leq f(Xi)$ and $\delta > (nn/N)$, i.e. position Xn has a better food consistency than the current position of AF i and population density in its neighborhood is not much, therefore AF i moves one step toward AF by Eq (7):

$$\vec{X}_i(t+1) = \vec{X}_i(t) + \frac{\vec{X}_n - \vec{X}_i(t)}{Dis_{i,n}} \times Step \times Rand(0,1) \quad (7)$$

If AF i has no neighbors or none of its neighbors satisfy following condition, prey behavior would be performed for AF i.

AFSA performs the optimization process iteratively by means of functions described as its behaviors, and the algorithm factors or AFs try to get closer to our objectives by executing these functions.

For AFs, prey and free-move behaviors are the individual ones and follow and swarm behaviors are group behaviors. Prey behavior would be performed if an AF can't move to a better location by executing follow behavior and/or swarm behavior, and free-move behavior is performed if an AF is not successful in finding a better location by performing prey behavior. Certainly, prey and free-move behaviors are not performed individually by AFs and are only performed when an AF couldn't move to a better location by follow and swarm behaviors. In AFSA, in each step of the optimization process, all AFs pass the same procedure in parallel.

Let the position of AF i at time t be $X_i(t)$. AF i performs follow behavior at position $X_i(t)$ once, which leads to obtain position $X_{i,Follow}$. After executing the follow behavior, AF i performs swarm behavior from that position $X_i(t)$ and it leads to obtain position $X_{i,Swarm}$. After performing both follow and swarm behaviors that were done with respect to $X_i(t)$, the next position of AF i ($X_i(t+1)$) is obtained by Eq (8):

$$X_i(t+1) = \begin{cases} X_{i,follow} & if \quad f(X_{i,follow}) \leq f(X_{i,swarm}) \\ X_{i,swarm} & if \quad f(X_{i,follow}) > f(X_{i,swarm}) \end{cases} \quad (8)$$

In AFSA, a bulletin is used for recording the best position that has been found so far by all swarm members. In each of the iterations, after performing AFSA's behaviors by all AFs and moving them to new locations, the fitness value of the best AF is compared with the recorded location on the bulletin. If fitness value of the best AF is better than the recorded location on the bulletin, the location of the best AF is recorded as the best found location so far. Standard AFSA pseudo code is shown in Figure 2.

```
Standard AFSA
foreach AF i
   initialize xi
endfor
bulletin = arg min f(Xi)
              Xi
Repeat
   foreach AF i
      Perform swarm behavior on Xi(t) and obtain Xi,swarm
      Perform Follow behavior on Xi(t) and obtain Xi,follow
         if f(Xi,swarm) ≥ f(Xi,follow) then
            Xi(t+1) = Xi,swarm ;
         else
            Xi(t+1) = Xi,follow ;
         endif
   endfor
   if f(XBest-AF) ≤ f(bulletin) then
      bulletin = XBest-AF ;
   endif
until stopping criterion is met
```

Figure 2. Pseudo Code of AFSA Algorithm

In this pseudo code, prey and free-move behaviors were considered as a part of swarm and follow behaviors. That is, if an AF could not perform successfully a swarm or follow behavior, it would perform prey behavior and if could not reach a better position by executing this behavior, it would perform free move behavior.

## 3. THE PROPOSED METHOD

In this section, the proposed clustering algorithm is described. In this algorithm, there exists a population of fish, where each fish includes n positions of clusters' center. If the clustering samples observe d-dimensional space, each artificial fish includes a position vector with n*d dimensions.

In the proposed method, in order to establish a balance between exploration and exploitation, some modifications have been applied to the structure of visual and step parameters. For this reason, Euclidean distance between all artificial fish and the best position, which is stored in bulletin, is calculated. Subsequently, the value of visual parameter for each fish is equal to v% of its distance to the best found position and the amount of step parameter for each fish is equal to s% of visual value. Based on the structure of swarm intelligent methods regarding convergence to the goal and reduction in swarm diversity by approaching the goal, Step and Visual parameters are reduced by approaching the goal. This issue leads to a balance between local and global searches, and consequently increases the efficiency of the algorithm.

Concerning the calculation process of visual and step parameters in the proposed method, each fish can use its visual and step parameters by equations 2, 3, 4, 6 and 7. It is worth to mention that regarding the difference between values of visual parameter for different fish, crowed factor and other related conditions are eliminated from follow and swarm behaviors of standard AFSA. By eliminating this parameter, convergence speed is increased in the proposed method which leads to premature convergence problem. To master this weakness, a novel mechanism is added to the algorithm.

In the new mechanism, each artificial fish moves with probability of p1%, based on the stored position in bulletin by eq. 9.

$$\vec{X}_i(t+1) = \vec{X}_i(t) + \left(\left(\vec{X}_{Bulletin} - \vec{X}_i(t)\right) \times Rand(-1,1)\right) \qquad (9)$$

where, $X_{bulletin}$ is equal to the best found position by the algorithm and Rand function generates a random number in range of [-1,1] with uniform distribution. According to Eq. 9, in case of generation the random number in range of [-1,0), the new position of the artificial fish is away from bulletin position, which leads to explorer the distant regions by the artificial fish and escape from local minima. On the other hand, in case of generation the random number in range of (0,1], the positions of the fish approaches the bulletin which leads to increase convergence speed. So, using Eq. 9, a balance between density and diversity is obvious, which leads to establish a balance between exploration and exploitation.

In addition, another mechanism has been considered in the proposed method in order to increase convergence speed in clusters. As it was mentioned before, each artificial fish included n × d-dimensional cluster centers. Since the nodes' clustering performs based on their position, and each node consists two components (X and Y geographical coordinates), the problem space for AFSA algorithm is a 2×n dimensional space. Position vector for each artificial fish is shown in Figure 3. Here $Z_{i,j}$ represents the dimension j from cluster head i.

$$(Z_{1,1}, Z_{1,2}, Z_{2,1}, Z_{2,2}, ..., Z_{n,1}, Z_{n,2})$$

Figure 3. Structure of position vector for solving node clustering problem in wireless sensor networks.

Fitness function that has been used for each artificial fish in the proposed method is the total of intra-cluster distances which is equal to the whole Euclidean distances between nodes and the nearest cluster head.

In the third mechanism, one cluster head changes its position with the probability of P2% in each iteration. This cluster head is arbitrarily chosen form entire set of cluster heads. To change the position of the selected cluster, first, all nodes that their distance to the selected head is lower than that to other ones are determined. Then, the new position of this head is equal to the center position of the determined nodes. In other word, the center is equal to the average position of the determined nodes.

| | Proposed Algorithm |
|---|---|
| 1: | Initialize WSN |
| 2: | **Do** |
| 3: | Initialize *AFSA* based on *5%* of alive node as *Cluster Number* |
| 4: | **Do** |
| 5: | foreach *AF* i |
| | /*After each movement, compare new position with Bulletin*/ |
| 6: | Compute *Visual* and *Step* for *AF* i |
| 7: | Perform swarm behavior on $X_i(t)$ and obtain $X_{i,swarm}$ |
| 8 | Perform Follow behavior on $X_i(t)$ and obtain $X_{i,follow}$ |
| 9: | **if** $f(X_{i,swarm}) \geq f(X_{i,follow})$ **then** |
| 10: | $X_i(t+1) = X_{i,swarm}$ ; |
| 11: | **Else** |
| 12: | $X_i(t+1) = X_{i,follow}$ ; |
| 13: | **Endif** |
| 14: | /*executing $1^{st}$ mechanism |
| 15: | **if** generated_rand_number < P1 **then** |
| 16: | Perform Eq.(9) |
| 17: | **Endif** |
| 18: | /*executing $2^{nd}$ mechanism |
| 19: | **if** generated_rand_number < *P2* **then** |
| 20: | Choose a cluster-head position randomly in X(t) and update its position based on cluster center |
| 21: | **Endif** |
| 22: | **Endfor** |
| 23: | **While** Stopping Criteria is met |
| 24: | Send Optimum Cluster Head Positions to WSN |
| 25: | Reduce Related Energies from All Nodes |
| 26: | **While** the Last Node Die |

Figure 4. Pseudocode of the proposed algorithm.

As it was mentioned, the considered fitness function for the proposed method is the total of intra-cluster distances. Nevertheless, the clustering process in wireless sensor network is a discrete process, since the cluster heads are selected among nodes. But, the proposed AFSA algorithm, similar to as most swarm intelligence methods, is used in continuous problems. To solve this problem, the movement of artificial fish in the algorithm is applied in form of continuous. However, the obtained continuous positions by the algorithm are assigned to the nearest node after each movement. It is worth to note that in order to prevent the death of nodes with low energy, first, the average energy of alive nodes are calculated and the nearest node that its energy is higher than the average energy is selected. Another important point is to determine the number of clusters. To determine this amount, we used the approach proposed in [22] i.e. 5% of the number of all alive nodes. Pseudocode of the proposed algorithm is shown in     Figure 4.

## 4. EXPERIMENTS

In this section efficiency of proposed algorithm for node clustering in wireless sensor networks is evaluated by experiments. First of all, configuration of the wireless sensor network which is used as a testbed for clustering problem is described. Next, comparative algorithms alongside their parameters' setting are explained. Lastly, the experiments results are examined.

### 4.1. Network Configuration

We considered a network involving 100 nodes which are distributed in a ground with the size of 100*100 meters, where the main station is placed in the coordinate of 50 and 175. The network is homogeneous (all nodes have the same energy) and it is considered that all nodes continuously includes the information that must be sent. Each node sends a packet toward the cluster head in each round and the head aggregates the gathered packet in the end of the round to be sent to the main station. The length of each packet is 500 bytes (4000 bits). It is considered that the main station is aware from the initial energy and positions of all nodes. Clustering algorithm is performed in the initial station with unlimited computational power and energy. In addition, we suppose that no information concerning the current energy is sent along with the information packet by the nodes and no redundant packet is sent in this regard.

### 4.2. Energy Model

In this research we use the simple radio model [22] in order to determine the radio energy. According to this model, for sending a message with k bit and the interval of d, the amount of energy which used by radio ($ET_x (k,d)$), is calculated based on Eq. 10.

$$ET_x (k,d) = E_{elec} * k + E_{amp} * k * d2 \quad (10)$$

where, $E_{elec}$ is the amount of consumed energy for establishing sender/receiver circuit. We consider the value of this parameter equals to 50 ($E_{elec}$=50 nJ/bit). $E_{amp}$ is the consumed energy by amplifier to amplification of the sent signal. We consider the value of this parameter equals to 100 ($E_{amp}$=100 pJ/bit/m2).

In addition, for receiving a message, the amount of energy which used by radio ($ER_x (k)$), is calculated based on Eq. 11.

$$ER_x (k) = E_{elec} * k \quad (11)$$

Cluster heads consumes more energy, since data aggregation is an additional phase for Cluster heads. We consider this value equals to 5 (EDA=5nJ/bit/signal).

International Journal of Computer Networks & Communications (IJCNC) Vol.7, No.3, May 2015So, the procedure of energy consumption in each round is as follow. First, all nodes receive a packet including the information for determining clusters and cluster heads (receive energy by nodes). Then, each node sends a packet to its cluster head (send energy by nodes and receive energy by cluster heads). Cluster heads aggregate the gathered packet in the end of each round to be sent to the main station (data aggregate energy and send energy by cluster heads).### 4.3. Swarm Intelligence Algorithms

In order to evaluate the efficiency of the swarm intelligence algorithms for the node clustering process, we have considered Particle Swarm Optimization (PSO) [23], Imperialist Competitive Algorithm (ICA) [24], Shuffled Frog Leaping Algorithm (SFLA) [4], Modified Artificial Fish Swarm Algorithm (CM-AFSA) and Leach algorithm [22] along with the proposed method. The First four algorithms belong to swarm intelligence algorithms category.

Particle Swarm Optimization (PSO) is the most recognized swarm intelligence algorithm proposed by Kennedy and Eberhart [2]. This algorithm simulates the fish and birds behaviors. Imperialistic cognitive Algorithm (ICA) was proposed in 2007 by Atashpaz and Lucas [24]. This algorithm is based on the social and political relations of imperialist and colonial countries. The shuffled frog-leaping algorithm (SFLA) that was developed by Eusuff and Lansey in 2003 [4], is a member of the swarm intelligence family. It is a meta-heuristic optimization technique based on the mimetic evolution of frogs seeking food in a pond. CM-AFSA is an enhanced version of AFSA algorithm in which PSO formulas are used to improve the performance of basic AFSA. In addition, a communication behaviour is applied in order to improve the efficiency of the algorithm.

For this reason, first, the number of clusters is considered equals to the 5% of the alive nodes [22]. Then, clustering process is performed by swarm intelligence algorithm and some mechanisms for mapping the center position of cluster heads to the nearest nodes with sufficient energy are applied for all algorithms. The algorithms continue the clustering process until reaching convergence condition. To determine the convergence condition, the central position of cluster heads in the current iteration is compared to that in 20 previous iterations. If there is no movement in the location of cluster heads for 20 iterations, convergence is executed.

The parameters' values for different algorithms are tabulated in Table 1. It is worth to note that the parameters are determined based on a wide range of experiments in this domain.

The experiments are iterated 50 times, where each iteration is performed up to 100 rounds in wireless sensor network. The positions of nodes are randomly determined in each iteration. The used criteria to evaluate the performance of the network are First Node Die (FND) and Last Node Die (LND). The average values of FND and LND are presented in table 2 for all tested algorithms. In addition, chart 5 illustrates the average remained energy in each round for different algorithms.

As can be seen in table 2, Leach algorithm led to the worst results, compared to other algorithms. However, it must be note that the computational cost of this algorithm is considerably lower than that of swarm intelligence algorithms. The results obtained by the swarm intelligence algorithms show that they led to different results, due to the performance of optimization process for clustering purpose. Among the swarm intelligence algorithms, the first node is died in a network which is clustered by ICA and PSO algorithms. The proposed algorithm outperforms other comparative ones in case of FND criterion and CM-AFSA algorithm led to the second best result.

Table 1: Parameters' values for different swarm intelligence algorithms for node clustering process.

| Algorithm | Parameter Setting |
|---|---|
| G-PSO | Population size =20×Cluster number<br>$c_1=c_2=2$<br>Inertia Weight=0.5×(rand*0.5)<br>neighbor topology=Global star |
| CM-AFSA | Population size =10×Cluster number<br>try-number=5<br>crowd factor=0.75<br>Visual=Par[1]<br>Step = Visual/5<br>$c_3=c_6=2$ |
| Std-ICA | Country number=20×Cluster number<br>empire number = 4×Cluster number<br>$\xi=0.1$<br>$\beta=2$ |
| SFLA | Population size = 10×Cluster number<br>number of memeplex=5<br>$D_{max}=20$<br>Iteration in memeplex = 10 |
| Proposed AFSA | Population size =10×Cluster number<br>try-number=5<br>Visual=Par[1]<br>Step = Visual/5<br>$c_3=c_6=2$<br>V% = 0.4 (40%)<br>S%=0.5(50%)<br>$P_1$%=0.3<br>$P_2$%=0.3 |

Table 2: Average and standard deviation of FND and LND by the algorithms

| Algorithm | FND | LND |
|---|---|---|
| Leach | 14.85 (0.74) | 21.05 (0.51) |
| G-PSO | 45.50(6.12) | 76.80(3.59) |
| CM-AFSA | 49.58(5.69) | 76.33(3.72) |
| Std-ICA | 45.60(5.20) | 76.20(4.64) |
| SFLA | 47.93(4.49) | 77.10 (4.18) |
| Proposed AFSA | 52.00(4.38) | 78.92(3.86) |

In addition, the proposed algorithm also outperforms other comparative ones in case of LND criterion and SFLA algorithm led to the second best result. It must be note that the nodes that are clustered using SFLA have died earlier than that using CM-AFSA, however, SFLA algorithm performed better clustering in the next rounds by decreasing the number of alive nodes. All the same, the proposed algorithms managed to achieve the best result both for FND and LND factors which illustrates the high performance of the proposed method in all series of the network. The reasons for such improvement, particularly versus CM-AFSA, are establishing a balance between local and global searches, increasing the escape ability from local minima and the high convergence speed. These matters are obtained thanks to novel mechanisms 1 and 2, as well as change in structure of visual and step parameters of artificial fish. Figure 5 shows the chart of total remained energy in nodes after 50 iterations by different algorithms.

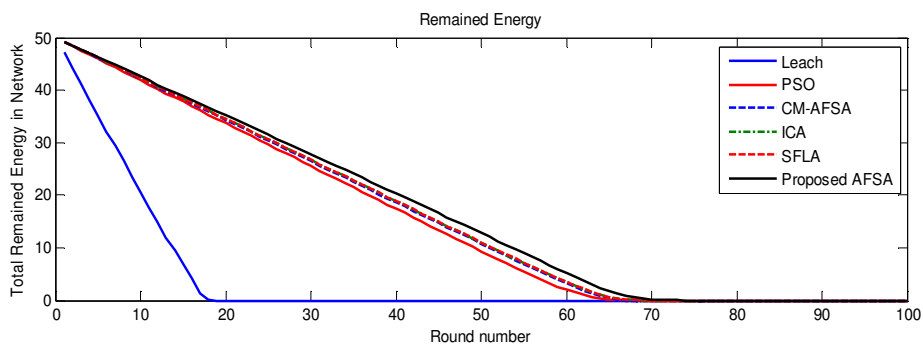

Figure 5. Energy reduction chart for all algorithms during 100 rounds.

## 5. CONCLUSIONS

In this paper, a novel artificial fish swarm algorithm was proposed. In the proposed algorithm, the value of visual parameter was determined by the distance of each fish from the best found position by the algorithm in each iteration. Even though, the preliminary search space of fish was huge, but it was decreased by convergence of the swarm. So, the searched space by each fish was commensurate with the progress of the algorithm and a balance between exploration and exploitation has been established. On the other hand, using two novel mechanisms i.e. movement based on the best found position and relocation of a cluster head toward the average position of the swarm's members led to improve convergence speed as well as escape ability form local minima. The experiments have been done for nodes' clustering in wireless sensor network and the performance of the proposed method was compared to several state-of-the-art algorithms in this domain. The results showed the superiority of the proposed method than comparative studies.

In this paper, a modified algorithm is used for clustering of nodes in wireless sensor networks. Since, there are many applications based on mobile wireless sensor networks in which nodes are able to move by time, node clustering in such networks can be a suitable topic for further researches. For this purpose, clustering algorithms must be redesigned to cover the dynamic environment of the problem.

## REFERENCES

[1] X.S. Yang, Z. Cui, R. Xiao, A.H. Gandomi, M. Karamanoglu, Swarm intelligence and bio-inspired computation: theory and applications, Elsevier, (2013) ISBN: 978-0-12-405163-8.
[2] J. Kennedy, R.C. Eberhart, Particle swarm optimization, IEEE International Conference on Neural Networks, 4 (1995) 1942-1948, DOI: 10.1109/ICNN.1995.488968.
[3] M. Dorigo, Learning and natrual algorithms, Ph.D. thesis, Dipartimento di Elettronica, Politecnico di Milano, Italy, 1992.
[4] M. Eusuff, K. Lansey, Optimization of water distribution network design using th shuffled frog leaping algorithm, Journal of Water Resources Planning and Management, 129 (2003) 210-225, DOI: 10.1061/(ASCE)0733-9496(2003)129:3(210)).

[5] L.X. Lei, Z.J. Shao, J.X. Qian, An optimizing method based on autonomous animate: fish swarm algorithm, System Engineering Theory and Practice, 11 (2002) 32-38.
[6] M. Neshat, G. Sepidnam, M. Sargolzaei, A. Nadjaran Toosi, Artificial fish swarm algorithm: a survey of the state-of-the-art, hybridization, combinatorial and indicative applications, Artificial Intelligence Review, (2012), DOI: 10.1007/s10462-012-9342-2.
[7] H.C. Tsai, Y. H. Lin, Modification of the fish swarm algorithm with particle swarm optimization formulation and communication behavior, Applied Soft Computing, 2011, DOI:10.1016/j.asoc.2011.05.022.
[8] W. Shen, X. Guo, C. Wu, D. Wu, Forecasting stock indices using radial basis function neural networks optimized by artificial fish swarm algorithm, Knowledge-Based Systems, 24 (2011) 378-385, DOI: 10.1016/j.knosys.2010.11.001.
[9] D. Yazdani, H. Nabizadeh, E.M. Kosari, A.N. Toosi, Color quantization using modified artificial fish swarm algorithm, Lecture Notes in Computer Science, 7106 (2011) 382-391, DOI: 10.1007/978-3-642-25832-9_39.
[10] D. Yazdani, M.R. Akbarzadeh-T, B. Nasiri, M.R. Meybodi, A new artificial fish swarm algorithm for dynamic optimization problems, IEEE Congress on Evolutionary Computation, CEC2012, (2012) 1-8, DOI: 10.1109/CEC.2012.6256169.
[11] G. Zheng, Z. Lin, A winner determination algorithm for combinatorial auctions based on hybrid artificial fish swarm algorithm, Physics Procedia, International Conference on Solid State Devices and Materials Science, 25 (2012) 1666-1670, DOI: 10.1016/j.phpro.2012.03.292.
[12] R. Azizi, Empirical study of artificial fish swarm algorithm, International Journal of Computing, Communications and Networking 3 (1) 1-7, DOI: arXiv:1405.4138 [cs.AI].
[13] A. Rocha, T. Martins, E. Fernandes, An augmented lagrangian fish swarm based method for global optimization, Journal of Computational and Applied Mathematics, 235 (2011) 4611-4620, DOI: 10.1016/j.cam.2010.04.020.
[14] D. Yazdani, A. Nadjaran Toosi, M.R. Meybodi, Fuzzy adaptive artificial fish swarm algorithm, AI 2010: Advances in Artificial Intelligence, Lecture Notes in Computer Science, 6464 (2011) 334-343, DOI: 10.1007/978-3-642-17432-2_34.
[15] D. Yazdani, S. Golyari, M.R. Meybodi, A new hybrid algorithm for optimization based on artificial fish swarm algorithm and cellular learning automata, 5th International Symposium on Telecommunications (IST), (2010) 914-919, DOI: 10.1109/ISTEL.2010.5734153.
[16] D. Yazdani, B. Saman, A. Sepas-Moghaddam, F.M. Kazemi, M.R. Meybodi, A new algorithm based on improved artificial fish swarm algorithm for data clustering, International Journal of Artificial Intelligence, 11 (2013) 193-221.
[17] J. Yick, B. Mukherjee, D. Ghosal, Wireless sensor network survey, Journal of Elsevier on Computer Networks, 52 (2008) 2292-2330, DOI: 10.1016/j.comnet.2008.04.002.
[18] Z. Rezaei , S. Mobininejad, Energy Saving in Wireless Sensor Networks, International Journal of Computer Science & Engineering Survey (IJCSES) Vol.3, No.1, February 2012, 23-37
[19] M. Parvin, E. Jafari, R. Azizi, A Multi-Armed Bandit Problem-Based Target Coverage Protocol for Wireless Sensor Network, Computing, Communication and Networking Technologies (ICCCNT), 2014 International Conference on , pp.1-5, 2014
[20] S. Getsy S,R. Kalaiarasi ,S. Neelavathy Pari, D.Sridharan, Energy Efficient Clustering and Routing in Mobile Wireless Sensor Network," International Journal of Wireless & Mobile Networks (IJWMN), vol.2, pp. 106-114, 2010
[21] A.A. Abbasi, M. Younis, A survey on clustering algorithms for wireless sensor networks, In Journal of Elsevier Computer Communication, 30 (2007) 2826-2841, DOI: 10.1016/j.comcom.2007.05.024.
[22] W.B. Heinzelman, A.P. Chandrakasan, H. Balakrishnan, An application-specific protocol architecture for wireless microsensor networks, IEEE Transactions on Wireless Communications, 1 (2002) 660-670, DOI: 10.1109/TWC.2002.804190.
[23] Y. Shi, R. C. Eberhart, A modified particle swarm optimizer, IEEE world congress on computational intelligence, (1998) 69-73 , DOI: 10.1109/ICEC.1998.699146.
[24] E. Atashpaz-Gargari, C. Lucas, Imperialist competitive algorithm: an algorithm for optimization inspired by imperialistic competition, IEEE Congress on Evolutionary computation, (2007) 4661-4667, DOI: 10.1109/CEC.2007.4425083.

**Authors**

**Reza Azizi** was born in Bojnourd, Iran in 1983. He received the BS degree in Software Engineering from Islamic Azad University of Shirvan, Iran, a MS degree in Computer Engineering from Eastern Mediterranean University, Cyprus. He has been employed as a research assistant at computer engineering department, where he has been involved in multiple projects. He is currently working as Lecturer at Educational Center of Applied-Science and Technology and a joint member of Islamic Azad University of Bojnourd. Research interests: Simulation, Performance Evaluation and Optimization of Wireless Sensor and Mobile Ad-Hoc Networks.

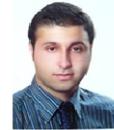

**Hamid Shoja** received the M.Sc. degree in Computer Software Engineering from the University of Tehran (PNU), Iran, in 2012. Since 2012 he has been worked as a researcher, system analyst and visualization manager at the Telecommunication Company of Iran, Bojnourd. In 2014, he joined Hakiman University of Bojnourd as a Faculty Member. His main areas of interest include visualization, load balancing issues in cloud computing and ad hoc wireless networks.

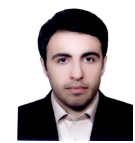